# Neuromorphic Optical Tracking and Imaging of Randomly Moving Targets through Strongly Scattering Media

Ning Zhang,[1,*] Timothy Shea,[2] and Arto Nurmikko [1,*]

[1] *School of Engineering, Brown University, 184 Hope St, Providence, RI, 02912, USA*
[2] *Intel Laboratories, Sunnyvale, CA, 94086, USA*
*[*ning_zhang1@brown.edu, arto_nurmikko@brown.edu](mailto:ning_zhang1@brown.edu)*

## ABSTRACT

Tracking and acquiring simultaneous optical images of randomly moving targets obscured by scattering media remains a challenging problem of importance to many applications that require precise object localization and identification. In this work, we develop an end-to-end neuromorphic optical engineering and computational approach to demonstrate how to track and image normally invisible objects by combining an event detecting camera with a multistage neuromorphic deep learning strategy. Photons emerging from dense scattering media are detected by the event camera and converted to pixel-wise asynchronized spike trains - a first step in isolating object-specific information from the dominant uninformative background. Spiking data is fed into a deep spiking neural network (SNN) engine where object tracking and image reconstruction are performed by two separate yet interconnected modules running in parallel in discrete time steps over the event duration. Through benchtop experiments, we demonstrate tracking and imaging randomly moving objects in dense turbid media as well as image reconstruction of spatially stationary but optically dynamic objects. Standardized character sets serve as representative proxies for geometrically complex objects, underscoring the method's generality. The results highlight the advantages of a fully neuromorphic approach in meeting a major imaging technology with high computational efficiency and low power consumption.

# INTRODUCTION

Tracking and identifying targets whose visibility is obscured and confounded by ambient light-scattering media presents a fundamental optical engineering challenge of importance for many applications. Examples range from biomedical imaging through opaque tissue, identifying vehicles in dense fog, free-space transmission of data carrying structured light through clouds, to many optical metrologies. Much work has been conducted to enhance the identification of *stationary* targets in turbid media by combining innovative optical methods with new computational techniques. These efforts feature applications of time-of-flight (ToF) and other time-gated approaches[1–4], laser speckle spectroscopy[5–10], and other photon correlation techniques[11,12]. In terms of optical hardware, advances in single photon avalanche detectors/cameras[13–15], LIDAR modules[2,16–18], new fluorescent methods[19,20] as well as optical engineering approaches such as wavefront shaping[21,22], ghost imaging[23], transforming multiple scattering to complex phase plates [24], have all made important contributions to the field. Taking advantage of advanced computational methods, today an integral part of any object recognition toolkit, researchers have combined a SPAD camera and a probabilistic ToF algorithm in an approach to 'photography' using a convolutional neural network model in scattering media[17] and introduced time-gated holographic techniques with coherent processing in fog and turbid water[25]. Elsewhere a group developed a stochastic optical scattering localization imaging technique approach to decode speckle correlations in super-resolution bioimaging[26]. Very recently, important new work has been reported which, while addressing a different problem of imaging in dynamic scattering media, develops an elegant learning-based approach to real-time imaging [27].

Nonetheless, the challenge of recognizing *randomly moving* objects in highly scattering media remains an open problem, requiring the ability to simultaneously track *and* reconstruct images in real-time. Several new research works have explored speckle and its correlation to track and image moving objects[5,8–10,28], in simple scattering medium (multiple scattering or ground glass) with reliance on structured motion or prior knowledge, sensitivity to object sparsity, and constraints imposed by the optical memory effect. As an example, we note a recent work[7] where a method of cross-correlating time series of speckle patterns was developed to classify and track but not image simple geometrical shapes.

In this paper, we make a leap forward by developing a fully neuromorphic strategy that integrates two key elements, namely event-driven sensing and neuromorphic computing to *both track and achieve full image reconstruction* of dynamical targets which are indistinguishable to the naked eye or conventional cameras. In developing the method, we have used the primate visual system as an inspiration, such as the symbiotic ability of the retina and the brain to jointly track and identify moving objects at very low light levels. Here the retina is a biomimicking dynamic vision sensor (DVS), a class of cameras of increasing application importance[28–31]. The computational brain in our work is modeled as a spiking neural network (SNN) in the approach that uses discrete spikes or "action potentials" by artificial neurons as the medium of information. Unlike other artificial neural networks (ANNs) in machine learning where neurons communicate through continuous-valued activations, the design and implementation of spiking neural networks for our purposes is attractive because of the sparse, event-driven nature which leads to energy efficiency[32,33] while being naturally compatible with emerging neuromorphic hardware. More generally, SNNs are well-equipped to handle spatiotemporal data in event-based scenarios in diverse applications such as pattern recognition[34], and bio-signal decoding[35].

In the growing field of spike-based neuromorphic computing and engineering[36,37], deep learning approaches have shown utility when dealing with high-dimensional problems associated with computer vision, for example to enhance robotic navigation in crowded environments[38]. For the work in this paper, we developed a multi-module deep SNN model with memory ability. This is a logical choice given our goal of preserving the event-driven spikes generated by the DVS sensor as the currency of information throughout the imaging system. From the viewpoint of minimizing system energy consumption and reconstruction latency, computing with spikes can be argued to offer advantages derived from intrinsic spatiotemporal sparsity. In both brain circuits and current deep learning accelerators (e.g. CPUs, GPUs), the most energy-intensive operation is the transmission of signals from pre-synaptic neurons to post-synaptic neurons – achieved through axonal propagation and matrix-matrix multiplication, respectively. Further, traditional signal processing requires that all channels (here frame-based camera pixels) comply with a global sampling rate. Since in most non-trivial imaging conditions activation of channels is uneven, this requires a compromise between running the system more slowly to allow the least active channels time to accumulate versus running more quickly to reduce the delay before processing the most active channels. By contrast, in event-driven asynchronous processing, there is no forced global synchrony in the transmission, hence lower latency as long as sparsity holds. It is of course well known that the visual system in primates and many animals is remarkably adept at detecting change being particularly sensitive to moving objects. Rather than continuously reconstructing static elements, the eye-brain symbiosis effectively filters out unchanging background scenery thereby conserving cognitive resources i.e. focusing only on newly introduced information. This selective attention to change contributes to the extraordinary efficiency of human perception. Inspired by this naturally optimized process, our work aims to emulate and leverage these principles.

A high-level overview of the end-to-end neuromorphic system approach in this paper is shown in Fig. 1. The retina-biomimicking silicon DVS sensor chip collects photons exiting the turbid media. Each camera pixel outputs an asynchronous spike train with a firing rate proportional to the dynamical changes within the object scene. We reasoned that under conditions of dense scattering, a DVS camera offers a hardware-based attention mechanism that could be effective in extracting meaningful information by favoring those photons that have interacted with a moving (or generally dynamic) target, thereby reducing the dominant obscuring and uninformative background. The spiking data from the camera feed a neuromorphic deep spiking neural network model which we develop here for dynamical targets with the ability to both spatially track and recover clear images of objects hidden from view. The SNN architecture, described in detail below, efficiently processes both spatial and temporal features, making it generally compatible with any event-driven imaging or decoding task. In analogy to neurons in the brain, the benefits of this approach are maximized when the data being processed are sparse vectors or tensors abundant in zeroes. Since inputs to neurons in the brain and to artificial spiking neurons are multiplied with synaptic weights to accumulate in a given neuron state, it follows that if most input values are zero ('0'), most of the neurons will require no expensive communication.

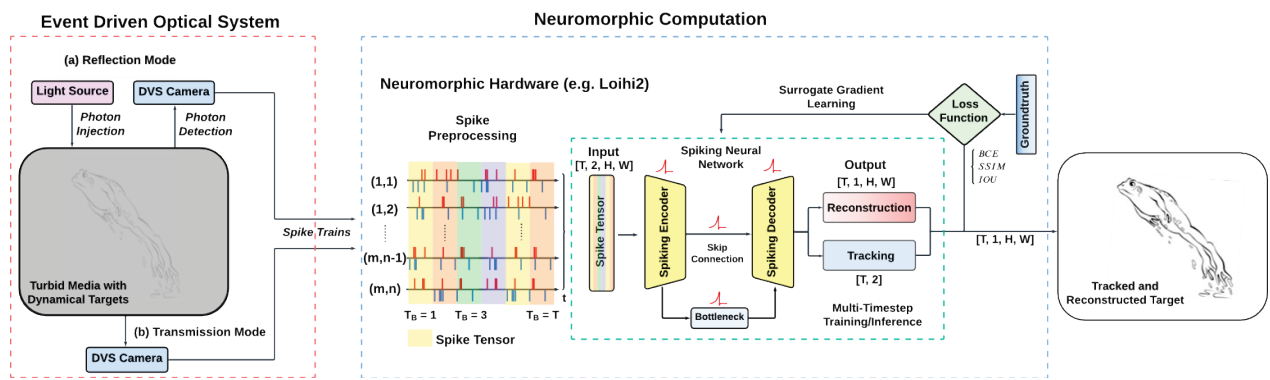

Fig 1. High level view of the end-to-end neuromorphic optical imaging approach for tracking and reconstructing images of dynamic targets obscured by turbid media (here cartoon of a jumping frog). The direction of data flow reflects the integration of the benchtop event-driven optical subsystem in either transmission of reflection mode (left) with the outline of the computational architecture of the deep spiking neural network (SNN) engine (right). Asynchronous spike trains generated by the optical subsystem are expected to favor those photons which interact with a dynamical target thereby reducing the processing of uninformative static background.

Below, we report on two types of benchtop experiments for object identification of dynamical optical targets through highly turbid media. In the first case the objects, images of two-dimensional characters chosen from MNIST dataset of handwritten digits are projected as structured light and move randomly in the transverse plane behind a slab of dense scattering media in transmission mode. In the second case, the objects (Kanji MNIST characters) were placed behind the turbid medium but now collected in reflection mode (hence doubling the optical pathlength). In the latter case, the character images are spatially stationary but made time varying in their optical intensity. The turbid medium is taken as static though could in principle be slowly varying in its optical density. The spike train output from a DVS sensor supplies the input data to the deep-learning-SNN model where two modules, one designed for tracking and the other performing image reconstruction, run in parallel across the event sequence such as during the random trajectory of the target.

As such, the field of neuromorphic computing is witnessing a veritable explosion in the literature including an array of hardware-tailored SNN models[39]; the reader is referred to recent reviews on the general subject such as by [37,40,41]. Many works have recently appeared where the combined benefits of an event-based sensor such as a DVS camera and dynamical computing with SNN algorithms operating in the deep learning regime have been recognized[42–44]. These have explored problems such as event-based trajectory prediction[45,46], object localization[47], optimization of SNNs[48], optical flow estimates[49], real-time speed detection[50], vision mimicking defocus model of depth [51], branching to applications such as 3D perception for stereo vision[52], intelligent driving[53], movement classification[54], gesture recognition[55], and others.. However, to the best of our knowledge, *in all optical detection/neuromorphic decoding* *work the ambient media is limited to optically transparent and unobstructed media*. We also mention work in our lab to investigate the use of SNN-based decoding of brain signals for movement intention in connection with bandwidth-efficient multichannel wireless transmission of data recorded from the motor cortex[56].

## RESULTS

### *EVENT-DRIVEN NEUROMORPHIC OPTICAL SYSTEM*

In a benchtop optical system (Figure 2) a series of generalizable proof-of-concept experiments was performed for both tracking and reconstructing simple 2D dynamical objects, which were obscured from view by a calibrated diffusive medium and indecipherable to the human eye. Experiments were conducted with (1) objects that randomly moved (in the x-y plane) in a transmission setup, and (2) objects that were spatially stationary but exhibited time-varying intensity in a reflection (backscattering) setup.

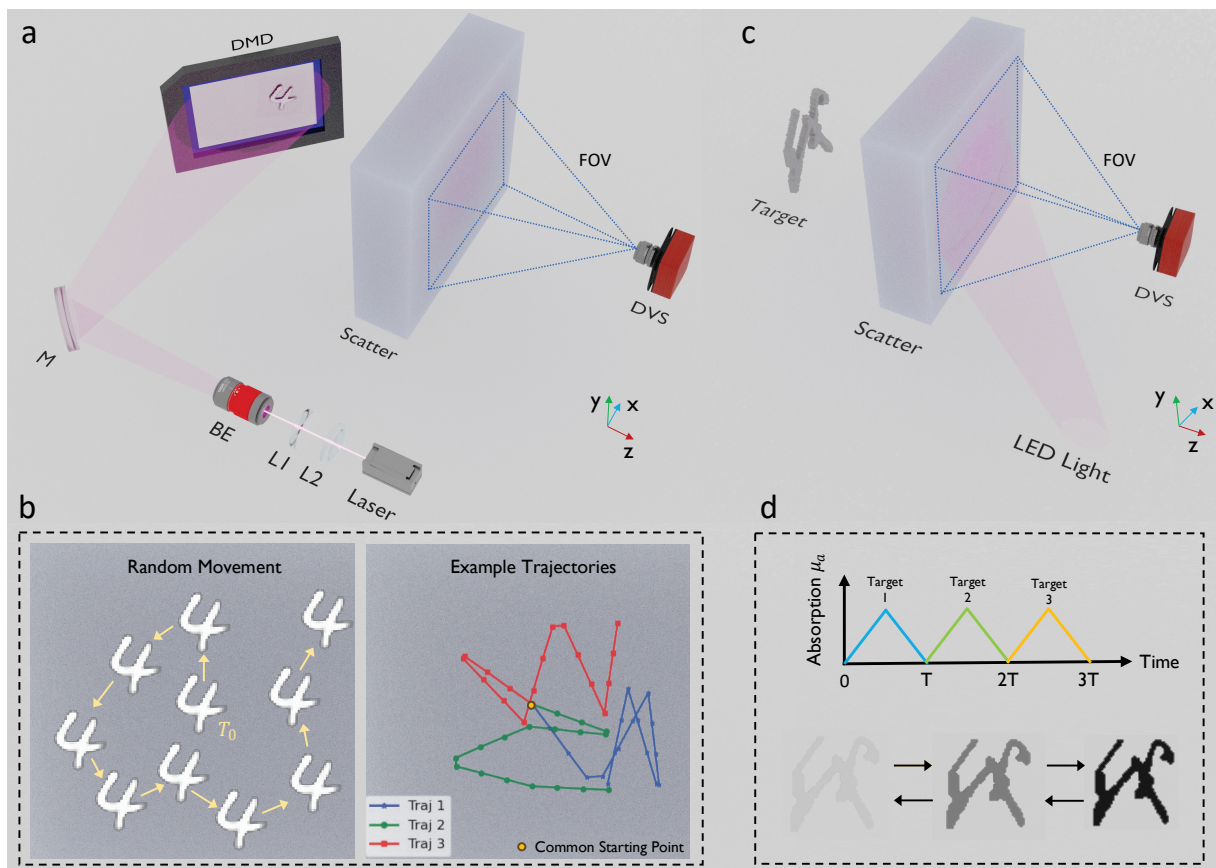

Figure 2. Optical setup for tracking and reconstruction of dynamical objects through scattering media. (a) Transmission geometry used for experiments with MNIST characters as objects. Structured light was projected via a digital micromirror device (DMD) to represent randomly moving objects. (b) Example of random trajectories for objects in the transmission setup, showing unpredictable, random patterns of movement in the x-y plane in time steps T. Left panel shows movement for a digit "4" while right panel displays the motion of the center of gravity of this character. (c) Reflectance (backscattering) setup used for experiments with Kanji-MNIST objects, utilizing an E-ink display to simulate spatially stationary objects with time-varying optical absorption property. (d) Representation of a Kanji-MNIST object appear and disappear periodically (absorption property changes) over time T, simulating dynamic targets.

In the transmission experiments, we employed a continuous-wave near-infrared laser (DL100, Toptica, λ= 808 nm, I =15 mW) as the illumination source. The beam, expanded through collimation, produced a spot with a radius of 15 mm, corresponding to an illumination fluence of 0.021 mW/mm². A digital micromirror device (DMD, Vialux Inc.) was used to scan and project structured light patterns in the x-y plane to simulate individual object trajectories moving randomly in the transverse plane relative to the optical axis, serving as a proxy for real physical targets. This technique of optical projection of images in lieu of 2D physical objects has been used by many authors[7,8,10,27,57–59] and does not subtract from the generalization of our approach. As shown in Figure 2b, each object's trajectory was programmed to be completely random (for both training and testing sets in deployment of the SNN-based deep learning model).

In the reflectance experiments, we used an LED light source for illumination (M850L3, Thorlabs, λ= 850 nm, I= 10 mW). To simulate spatially stationary objects with time-varying optical intensity contrast, an electronic ink display was utilized to gradually adjust the grayness of the target, based on characters from the Kanji MINST set. The working principle of E-ink technology, which relies on the movement of magnetic black and white particles, mimics time-varying optical absorption changes and is well suited for the purpose at hand. It has also been adopted to represent passive targets by other researchers[4,27,60]. Figure 2d shows schematically how each object appears and disappears once in a period T (in our work T was approximately 400 ms; speed limited by current E-ink technology).

As scattering media, we constructed silicone-based phantoms with microscale $SiO_2$ particles as Mie scatterers (see Methods). In the experiments, the scattering coefficient of the phantom was fixed at $\mu_s$ = 6 mm$^{-1}$ and anisotropy number as g=0.9 [16] so that for a slab thickness of 12 mm the resulting number of photon mean free paths (random scattering events) was MFP = 72 in the transmission geometry. In the reflection geometry we placed the target behind a densely scattering medium, to mimic applications that might involve the identification of a vehicle or pedestrian obscured by fog, or hemodynamically active vasculature in the brain for biomedical imaging. In each instance, a source and a detector locate on the same plane so that photons launched from the source must traverse the scattering medium twice before backscattered light is collected at the detector (thus MFP = 144).

As the event detector, we used a commercial DVS camera by Prophesee Inc (Gen3.0 EVK). Equipped with an active area of 9.6 × 7.2 mm2 with 640 × 480 pixels the camera includes a lens system mounted in front of the CMOS sensor chip to enhance the collection efficiency of photons. As noted, unlike traditional frame-based cameras, a DVS sensor generates a spike train only when the relative light intensity at a specific pixel is subject to a dynamical change. This key feature makes the device a highly efficient event-driven sensor with a high dynamic range of up to 140 dB [29] and a high temporal resolution in the order of microseconds (μs) per pixel, with added benefits of low energy consumption and minimal latency, significant assets for any real-time optical imaging system. We note that due to the stochastic nature of light scattering and photon detection by the event-based sensor electronics, there is no deterministic relationship between the original dynamic targets and the event sequences recorded by the DVS.

While configured on a benchtop in the laboratory for proof-of-concept demonstration, the simplicity of the optical scheme should allow implementation in field-portable or body-wearable use when deploying compact components such as a low-power semiconductor laser source, miniaturized optical lenses and other micro-optical elements.

*DEEP SPIKING NEURAL NETWORK MODEL (SNN)*

A particular challenge for SNNs is the non-differentiable nature of the various spiking neuron models which complicates the training process that use standard backpropagation techniques[61,62]. Common methods to train an SNN are: (1) Convert ANNs to SNNs[63], (2) Train with Local learning rules[64], and (3) deploy backpropagation with a surrogate gradient[61] adapted to the spiking domain. In this work, we chose the surrogate gradient method to train our SNN model (see Methods) primarily as this allows end-to-end training that makes effective use of temporal information in the input.

A standard leaky integrate-and-fire (LIF) neuron model was chosen to incorporate time-dependent dynamics and discrete spike outputs, mimicking biological neurons more closely than neurons in ANNs (see Methods). An LIF neuron maintains a membrane potential (an internal state, representing local learning and memory ability), exhibits a leakage of potential over time (representing local forgetting), and fires spikes when a threshold is reached followed by a refractory period.

***Designing the SNN Architecture***. Our goal was to design an algorithm specifically capable of both tracking and imaging dynamic objects obscured by scattering media. The input data reflects changes in the scattered photon population registered by the DVS camera. These changes may be small and accompanied by significant background noise. After the very first hardware layer of denoising and feature extraction in the DVS, the subsequent software model should be able to locate objects, discover and extract key representations out of the background to achieve sharp image reconstruction. More generally, the SNN must be able to efficiently process both spatial and temporal features for practical event-driven imaging and be deployable on neuromorphic hardware.

In an SNN model, the spiking information propagates as spikes and the output is reconstructed as tracked object images over a time series: $f_\omega(p(S_{(x,y,p,t)})) = R(T)$, where ω represents learnable parameters, $S_{(x,y,p,t)}$ is a spike train (x, y, polarities, time), $p()$ denotes pre-processing (see Methods) to convert spike trains into spiking tensor for training and inference, and R(T) denotes reconstructed images at time-step T.

We approached building our SNN engine by structuring it into two separate modules, each designed to address a specific task of object tracking and object reconstruction, respectively. Summary of the architecture is depicted in Figure 3. Given an input sequence $I(T, 2, H, W)$, the input branches simultaneously to the *SNN Reconstruction Module* and the *SNN Tracking Module*. At each time step ($T_i$, i = 1, …, N), the outputs from the two modules are combined for a joint tracking and reconstruction result.

*SNN Object Tracking Module (OTM)*

The primary objective of tracking is to estimate the center coordinates of a randomly moving tracked object at each time step (the center of the DMD projected image). The Object Tracking Module [OTM] in lower portion of Fig. 3 takes the incoming spikes and outputs the predicted coordinates [x, y] in the 2D plane (this work focused on 2D motion; extending the approach to depth perception is a goal of future studies). The OTM framework consists of two main components: an encoding path and a linear mapping path. The encoding path incrementally extracts critical spatial information and encodes it into a latent representation, while the linear mapping path transforms the flattened latent data into the desired coordinate format [x, y] through a series of fully connected layers.

The encoding path uses a sequence of convolutional SNN blocks, each followed by a downsampling step. We denote one such single-convolution SNN block as:

$$\boldsymbol{SNN\_Block} = \boldsymbol{SN}\left(BN(Conv2d(x))\right), \qquad [1]$$

where Conv2d(·) is a 2D convolution, BN(·) denotes batch normalization, and SN(·) is a spiking neuron operation (here LIF neurons). Starting from the input $x_0^* = x$, each encoded output $x_i^*$ is obtained via:

$$x_i^* = \overline{\boldsymbol{DS}}\left(\boldsymbol{SNN\_Block}(x_{i-1}^*)\right), \quad x_0^* = x; \ x_i^* \in \mathbb{R}^{T \times C_i \times \frac{H}{2^i} \times \frac{W}{2^i}}; \ i = 1, \dots, n \qquad [2]$$

Here, $\overline{\boldsymbol{DS}}$ is a downsampling operator (e.g., maxpooling) that halves the spatial dimensions (H, W) at each stage, and T is the number of time steps processed by the network.

Once the encoding path has generated a latent representation, the linear mapping path converts the flattened features into [x, y] coordinates. Each fully-connected (FC) SNN layer is given by:

$$x_j^* = \boldsymbol{SN}\left(FC(x_{j-1}^*)\right), \quad x_j^* \in \mathbb{R}^{T \times C_j}, \qquad j = 1, \dots, m \qquad [3]$$

where FC(·) is a fully connected layer. After the last block, the final output layer provides two neurons corresponding to the [x, y] coordinates normalized to the range [0,1].

The output is derived from the membrane potential of the last spiking neuron (SN) layer. At each time step $T_i$, the last SN layer accumulates spikes from the preceding layers to update its membrane potential. Notably, only the last SN layer undergoes resetting at each time step while the rest of the network remains unreset. This resetting strategy is employed for two primary reasons: (1) Handling random object movement: The tracked object may move unpredictably, sometimes transitioning from a position with a high normalized value to one with a low normalized value. While the SN layer effectively accumulates incoming spikes to increase its membrane potential, it is less efficient at rapidly decreasing the potential as needed. Although a constant potential decay occurs due to the leakage mechanism, this decay may be insufficient for substantial reductions in membrane potential. Without resetting the last SN layer, such scenarios could result in incorrect output estimations. (2) Preserving memory in intermediate layers: The remaining layers in the network do not require resetting because the network is trained over a series of time steps. The "stateful" parameters, such as the membrane potentials of intermediate SN layers, act as memory elements in the LIF model. These parameters capture contextual information from both previous and current positions, enabling the network to leverage temporal dependencies and achieve higher accuracy in predictions.

*SNN Object Reconstruction*

The SNN Reconstruction Module includes two sub-modules for final object reconstruction (upper portions of Fig. 3). The first, an Object Reconstruction Module (ORM) follows a U-Net-like [65] encoder-decoder structure which provides coarse spatiotemporal reconstruction. Inspired by BASNet design [66], a multiscale Residual Refinement Module (RRM) follows after the ORM to refine the resulting reconstruction by learning the residuals between the ORM reconstruction and the groundtruth.

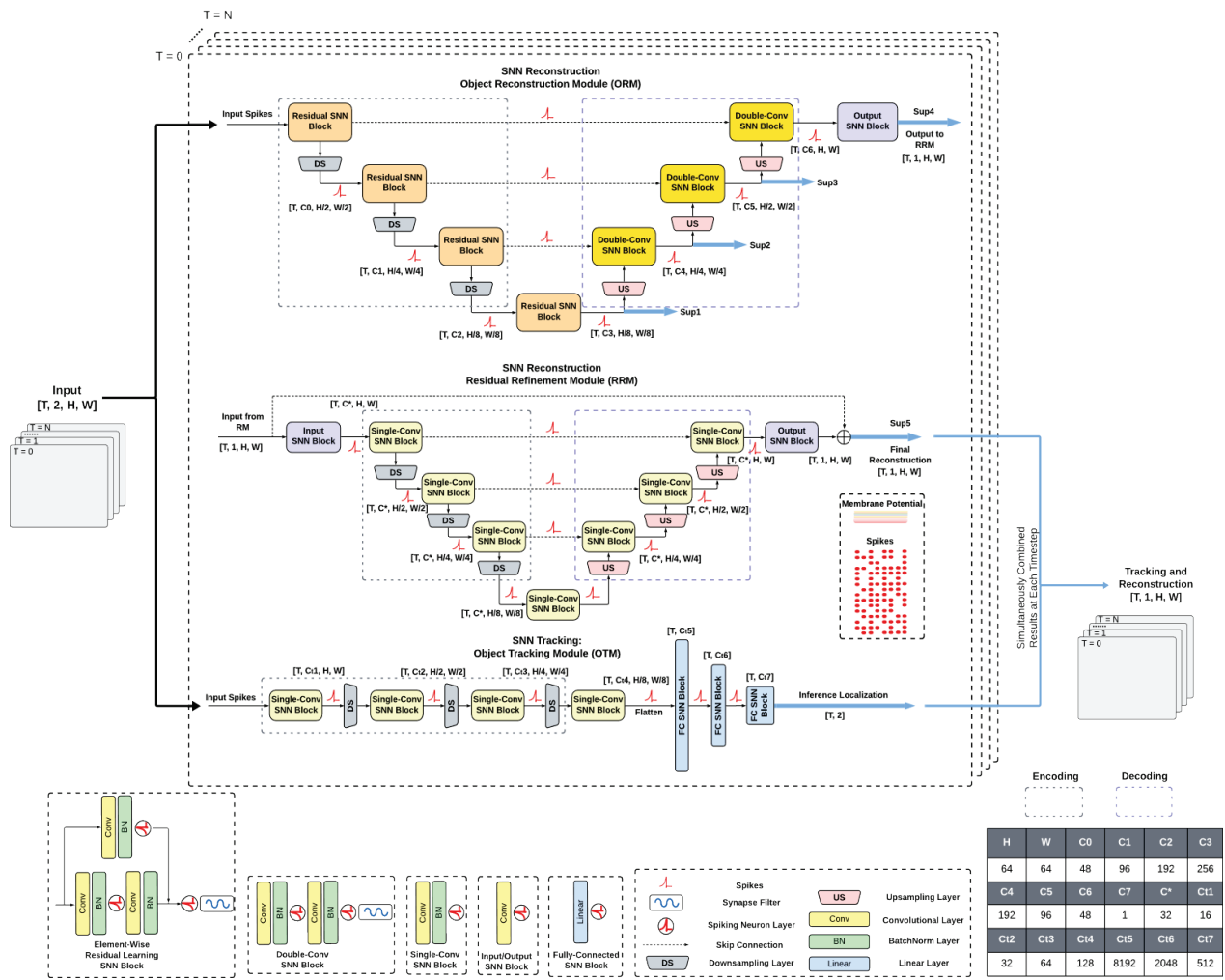

Figure 3. SNN Architecture for Dynamic Object Tracking and Reconstruction. The architecture is designed for event-driven imaging, utilizing a two-module approach to simultaneously track and reconstruct dynamic objects obscured by scattering media. The SNN model leverages spiking neural computation to process both spatial and temporal features. Input spike trains from the DVS are preprocessed into spiking tensors and processed through two distinct modules: Object Tracking Module (OTM) and Object Reconstruction Module (ORM). The SNN OTM in lower part of the large panel in Fig. 3 performs object localization, outputting normalized x-y coordinates of the center of the object at each time step. Its structure consists of an encoding path, which extracts spatial information into latent representations, and a linear mapping path which transforms this data into coordinate predictions. The x-y coordinates are derived from the membrane potential of the last spiking neuron (SN) layer which resets per time step to handle random object movements and preserve temporal context in the intermediate layers. The SNN ORM, based on a spiking U-Net-like framework in the uppermost panel, reconstructs images of objects from input spike trains. The module consists of an encoding path, a decoding path, and skip connections for retaining fine-grained spatial details. The encoding path progressively reduces spatial dimensions, capturing critical features in a bottleneck layer. The decoding path restores spatial resolution using upsampling layers and convolutional SNN blocks. The output from the reconstruction module inputs to a Residual Refinement Module (RRM) which follows the initial reconstruction and refines the final outputs by learning residual differences between coarse reconstructions and ground truth, improving boundary sharpness and regional consistency. Hybrid loss functions optimize training, with multi-stage losses applied from the bottleneck to each decoding stage (denoted as "SUP" in blue arrows). Reconstruction loss combines Binary Cross-Entropy (BCE) for pixel accuracy, Structural Similarity (SSIM) for patch coherence, and Intersection over Union (IoU) for map-level coverage. Tracking loss uses Mean Square Error (MSE) to refine predicted object center coordinates. Structure of the various SNN Blocks are shown at the bottom of the figure, including element-wise residual learning SNN blocks, stateful synapse filters, and others. The architecture incorporates nearest neighbor interpolation upsampling layers and max-pooling downsampling layers to efficiently manage spatial transformations. Spiking neurons in all layers encode spatiotemporal information into membrane potentials and spike outputs, aligning with the event-driven nature of the input data.

*Object Reconstruction Module (ORM)*

An encoder-decoder structure is chosen because it can computational efficiently capture both global context and local details while discarding noise and irrelevant details[65,66]. ORM consists an encoding path progressively downsampling the spatial dimensions of the input feature maps and a decoding path reconstructing the output at the original resolution, incorporating skip connections to preserve fine-grained spatial details from early stages. At the center of the network, where the downsampling reaches the deepest level, i.e. a bottleneck where the spatial dimensions are smallest, the model captures the most important features of the input spikes.

Mathematically, the ORM can be expressed as:

*For the i-th encoding blocks (Residual SNN Block with Downsampling):*

$$x_i^* = \overline{DS}[SN\left(BN\left(Conv2d(SNN\_Block(x_{i-1}^*))\right) + SNN\_Block(x_{i-1}^*)\right)],$$

$$x_0^* = x; \quad x_i^* \in \mathbb{R}^{T \times C_i \times \frac{H}{2^i} \times \frac{W}{2^i}}; i = 1, \dots, n. \qquad [4]$$

*For the j-th decoding blocks (Double-Conv SNN Block with Upsampling):*

$$y_j^* = SNN\_Block(SNN\_Block(\overline{US}(y_{j-1}^*) + SKIP)), \qquad [5]$$

$$SKIP = x_{n-j}^*; \; y_0^* = x_n^*; \; y_j^* \in \mathbb{R}^{T \times C_{j+n} \times \frac{H}{2^{n-j}} \times \frac{W}{2^{n-j}}}; \; j = 1, \dots, n.$$

*Output block:*

$$Output = SN(Conv2d(y_n^*)), \quad Output \in \mathbb{R}^{T \times 1 \times H \times W}; \; y_n^* \in \mathbb{R}^{T \times C_{2n} \times H \times W}$$

where $I \in \mathbb{R}^{T*C \times H \times W}$ is the input data after preprocessing, and $T, C, H, W$ denote the timestep, channels, height, and width, respectively. SKIP represents skip connection from the encoding blocks, concatenating the up-sampled spikes from the decoding path.

The encoding path consists of multiple *encoding blocks*. Each *encoding block* includes a residual SNN block and a downsampling (DS) layer, which progressively reduce the spatial resolution while increasing the number of feature channels. Each residual SNN block contains double-convolution layers, batch normalization (BN) layers, and spiking neuron layers SN(·).

In the decoding path, the network employs multiple *decoding blocks* which consist of upsampling (US) layers to gradually restore the spatial resolution. The US layers are interleaved with double-conv based SNN blocks to converge the reconstructed feature maps. Skip connections are utilized between corresponding layers in the encoder and decoder to fuse high-resolution features from the encoding path with the upsampled features in the decoding path.

In our method, the US layer is a bilinear layer or a transposed convolution layer. The DS layer is a maxpooling (MP) layer. The DS/US layer performs spatial reduction/restoration by a factor of 2.

The final output block includes a convolutional layer followed by a final SN layer. The convolutional layer transforms channels into 1 (gray image). This last SN layer processes the reconstructed spike events into the final output (membrane voltage or spikes) for training and inference.

A stateful synapse filter is added after each residual SNN block and double-conv SNN block. By placing it to filter the output current, the network can acquire stateful synapses which further enhance its memory ability[67].

We emphasize how spiking neuron layers SN(·) serve as a crucial and unique component in our architecture, efficiently encoding spatiotemporal information into membrane potentials and subsequently transforming them into binary spikes. This process enables spike-driven computation in subsequent layers, aligning with the event-based nature of our input data. Each SN layer, built from LIF neurons, takes spikes from the previous layer, updates its membrane potential (tensor U), and fires spikes (tensor S) as output: $S = SN(U)$.

*Residual Refinement Module (RRM)*

The reconstruction output from the ORM is a probability map, where each pixel value ranges between [0, 1], representing the likelihood of reconstruction at that location. However, the probability predictions from RM are prone to two common issues: (1) blurry and noisy boundaries, and (2) uneven regional probabilities[66]. To mitigate these problems, a Refined Reconstruction Module (RRM) is incorporated following RM.

The structure of the RRM is simpler than that of the ORM, consisting of an input block, three encoder blocks, a bridge, three decoder blocks, and an output block. The RRM is designed to learn the residual differences between the ORM's output and the ground truth, following the formula:

$$R_{final} = R_{coarse} + R_{residual} \qquad [6]$$

Where $R_{coarse}$ is coarse reconstruction generated by the ORM; $R_{residual}$ is output from the RRM prior to adding the input; $R_{final}$ is the final refined reconstruction produced by the model.

Our image reconstruction model demonstrates two key unique features: (1) Memory Capability: The model is built on a SNN instead of an ANN. Unlike ANN neurons, spiking neurons such as LIF neurons maintain internal states that store memory across time steps. This memory capability is further enhanced by the use of synapse filters, which transform synapses into stateful synapses. Consequently, both neurons and synapses in the model contribute to its memory retention. (2) Training Over Multiple Time Steps: The network is trained across multiple time steps rather than just a single time step, enabling it to handle the random spatial and temporal motion of obscured objects. Since a randomly moving object may appear in different locations at each time step, the model requires robust memory to contextualize the detected spiking patterns over time. This is crucial for addressing motion variance in DVS detection, where the same object can generate entirely different spiking patterns due to variations in its movement direction, acceleration, and other motion-related factors. Furthermore, different objects may exhibit similar spiking patterns under varying conditions. Thus the network's ability to retain memory of spiking patterns across time ensures it can learn and reconstruct objects effectively.

*Hybrid Loss Functions for Training*

In the SNN Reconstruction Module, the loss calculation is made at multiple stages (denoted as "SUP" in Fig. 3). At each SUP stage, the membrane potential from the last spiking layer is upscaled to match the input's shape and compared with the ground truth for loss calculation. This multi-stage loss strategy is implemented to mitigate overfitting and ensure robust learning[68]. A total of five loss calculations are performed at various stages, including outputs from both the ORM and RRM modules. To achieve accurate and high-quality reconstructions, globally and pixel-wise, a hybrid loss function is employed, comprising three components: (a) Binary Cross-Entropy (BCE) Loss[69] evaluates pixel-level accuracy by comparing each pixel in the reconstruction with the corresponding ground truth value. (b) Structural Similarity (SSIM) Loss[70] assesses reconstruction quality at the patch level, ensuring structural coherence and perceptual similarity to the ground truth. (c) Intersection over Union (IoU) Loss[71] measures overlap at the map level, encouraging better overall reconstruction coverage and accuracy.

Within the SNN Tracking Module, the loss calculation involves comparing the predicted center coordinates with the ground truth center coordinates. The Mean Square Error (MSE) is used as the loss function to minimize the error between predicted and true center positions, facilitating precise tracking (details in Methods).

## TRACKING AND RECONSTRUCTION – CONVERGENC OF THE EXPERIMENT AND SNN ALGORITHM

Using the benchtop optical setup of Fig. 2, we conducted a series of experiments in transmission and reflection (backscattering) geometries, respectively, to evaluate the performance of our neuromorphic optical imaging approach in dense scattering media. Results in the transmission mode of tracking and reconstructing randomly moving targets using MNIST digits as a proxy for 2-dimensional dynamic objects are shown next (As already noted, this technique of optical projection of images in lieu of 2D physical objects has been used by many authors[7,8,10,27,57–59] and does not subtract from the generalization of our approach), followed by image reconstruction results in the refection mode of spatially stationary Kanji-MNIST characters with time-varying optical contrast.

Figure 4 and the Supplemental video show an example of an MNIST digit '4' moving randomly in the x-y plane with the object placed behind the thick scattering medium (structured light by a DMD projector at 100 Hz). Panel (a) illustrates the raw spikes recorded across some 500 DVS camera pixels displayed as an x-y-t space-time graph. Note the presence of both positive and negative spikes from increase and decrease of transient illumination, respectively. The raw spikes are preprocessed (see Methods) into spiking tensors, shown in upper part of panel (b), represented in the format *[T, C, H, W]*, where *T* labels the time steps, *C* is the number of channels, and *H, W* correspond to the spatial dimensions. Object tracking by the OTM (lower part of panel (b)) is shown as inferred object position (green crosses) when compared against the ground truth positions (red crosses). Panel (c) illustrates the reconstructed images (integrated outputs of the OTM and ORM modules) from time steps T1 to T18. During the initial time steps (T1 to T6), the reconstruction appears blurred and lacks discernible shapes. Such behavior is reminiscent of the human visual process where temporal memory plays a role in object recognition such as the brain building a recognizable image of a moving target in conditions of weak ambient light. Here, the SNN algorithm leverages temporal memory, utilizing early-stage input data to progressively refine its understanding and recognition of the object. Across multiple time steps, the reconstruction process converges, with the object becoming increasingly clear and well-defined.

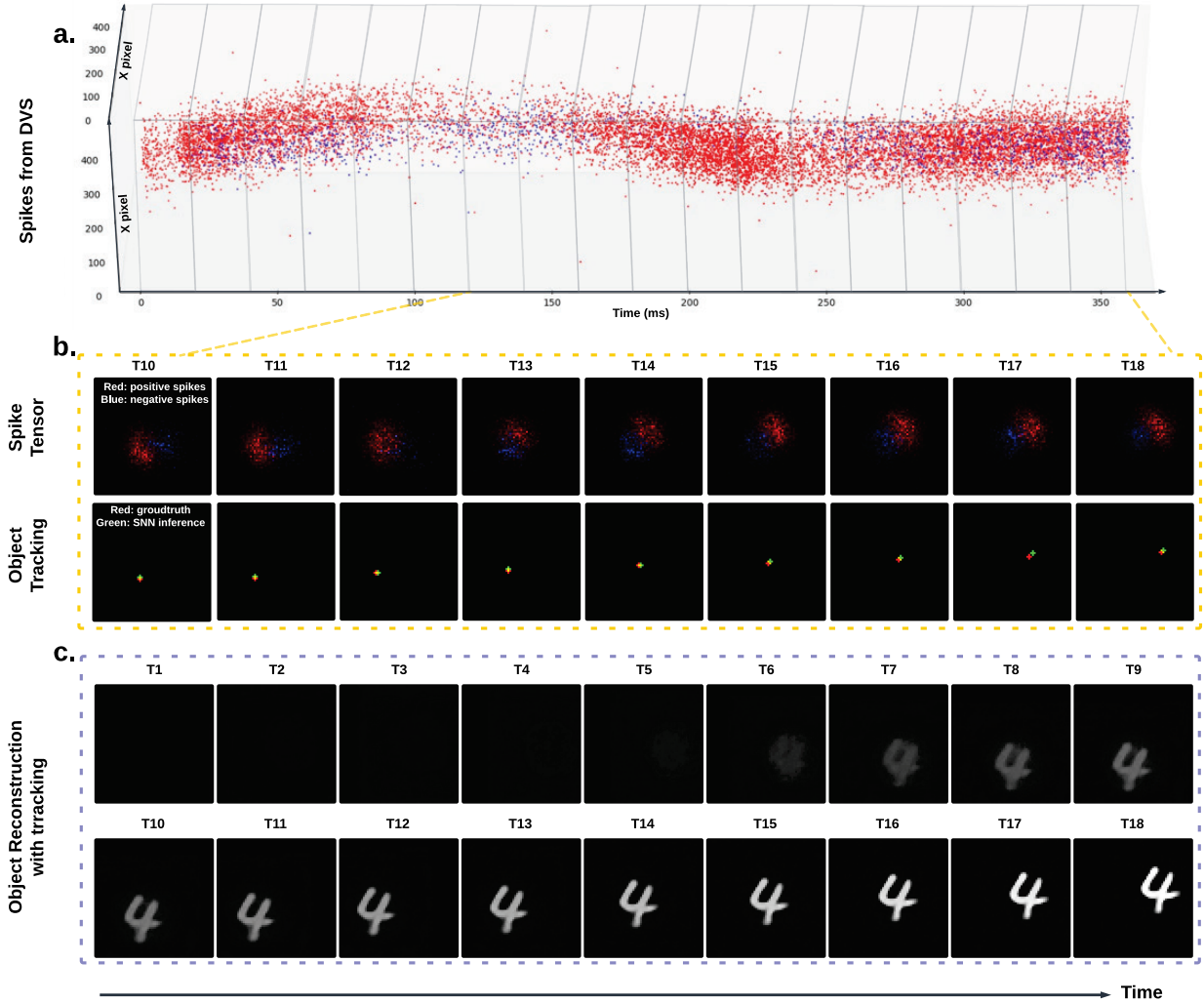

Figure 4. Neuromorphic image reconstruction of a moving object invisible to naked eye in in dense scattering media. (a, b) Data and results for an MNIST digit '4' moving randomly within the scattering medium as an x,y,t plot. (a) Sample of raw spiking data from a grpup of some 500 DVS pixels. (b) Spiking tensor derived from the preprocessed spikes, represented in the form [T, C, H, W] in upper panel. Object tracking, obtained from the OTM, illustrates the spiking neural network (SNN) inference (green crosses) alongside the ground truth positions (red crosses) of the randomly moving object in the lower panel. (c) Object reconstruction with tracking is the combined results from OTM and ORM at each time step. During a total 18 time steps the image of the object emerges as clear and fully recognizable (snapshots from a video in thre Supplement). The average structural similarity index measure (SSIM) and mean square error (MSE) between reconstruction and groundtruth is 0.9568 and 0.0108, repectively.

Figure 5 summarizes the reconstruction results obtained in transmission geometry for 64 randomly selected objects from the MNIST set, aggregated here as an 8×8 array where each image represents a distinct object which executed a trajectory of random motion for the same event duration (epoch). The first 10,000 characters from the full training set and 2,000 characters from the testing set collected from the optical experiments were used for training and testing, respectively. Using a standard computer equipped with an NVIDIA RTX 3090 GPU, training typically required approximately 6 minutes per epoch, while inference time averaged 15 ms for all time steps of a single target. This performance is expected to improve significantly with the adoption of dedicated neuromorphic hardware. Figure 5 shows a direct comparison between the raw unrecognizable spiking data recorded by the DVS camera through the phantom (optical thickness MFP = 72 scattering events), the "ground truth" data, and the reconstruction results generated by the SNN engine, respectively. With a structural similarity index measure (SSIM) of 0.9577 and a mean squared error (MSE) of 0.0108 across the full testing dataset, these results provide evidence supporting the effectiveness of the neuromorphic imaging strategy. The "live" reconstructions for both transmission and reflection geometries can be found in the supplementary section in time-lapse video format.

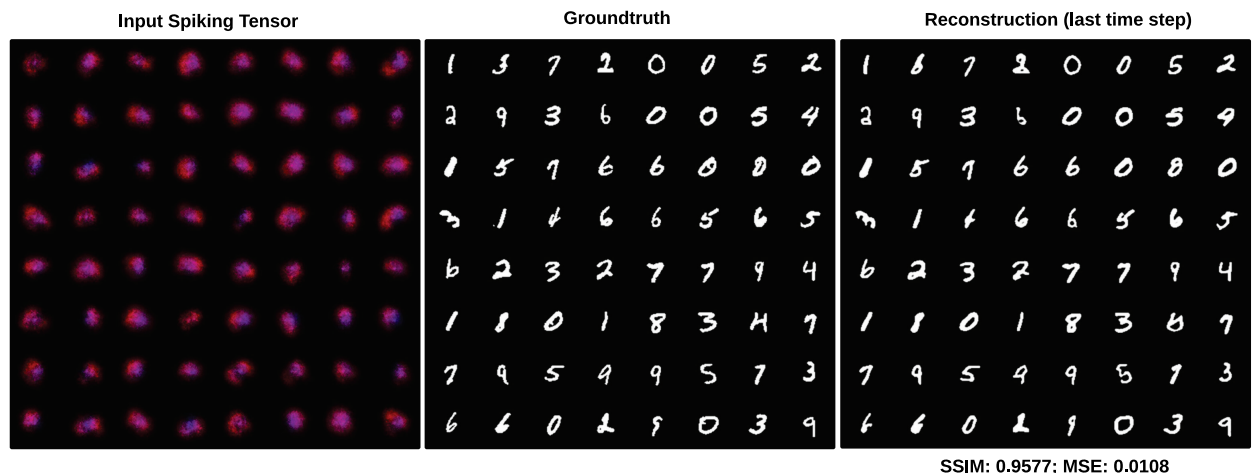

Figure 5. Summary of reconstruction results obtained in transmission geometry for randomly selected examples from MNIST, displayed as an 8×8 array where each image represents a distinct object. The figure compares the "ground truth" data, the raw spiking data recorded by the DVS camera through the phantom (MFP = 72), and the final SNN-generated reconstruction. The input image array shows the temporally accumulated spike tensor across all time steps, reflecting the random motion trajectories and unrecognizable blurriness caused by the scattering media. The ground truth image array displays the spatial information of the objects, while the reconstructed image array shows the reconstructed target at the final time step. Quantitative evaluation yields an SSIM of 0.9577 and an MSE of 0.0108, demonstrating the efficacy of this approach. Note, the size of objects in ground truth and reconstruction image are magnified for visualization purposes. Time-lapse videos of image reconstructions are shown in Supplementary section.

Figure 6 presents the results of experiments conducted in reflection geometry using Kanji characters as spatially fixed but optically time-varying objects. Each character was dynamically presented—emerging and disappearing (on and off only once per object)— using the E-ink display behind the scattering phantom. The problem becomes significantly more challenging due to the optical double-pass through the phantom (total MFP = 144), where opacity scales

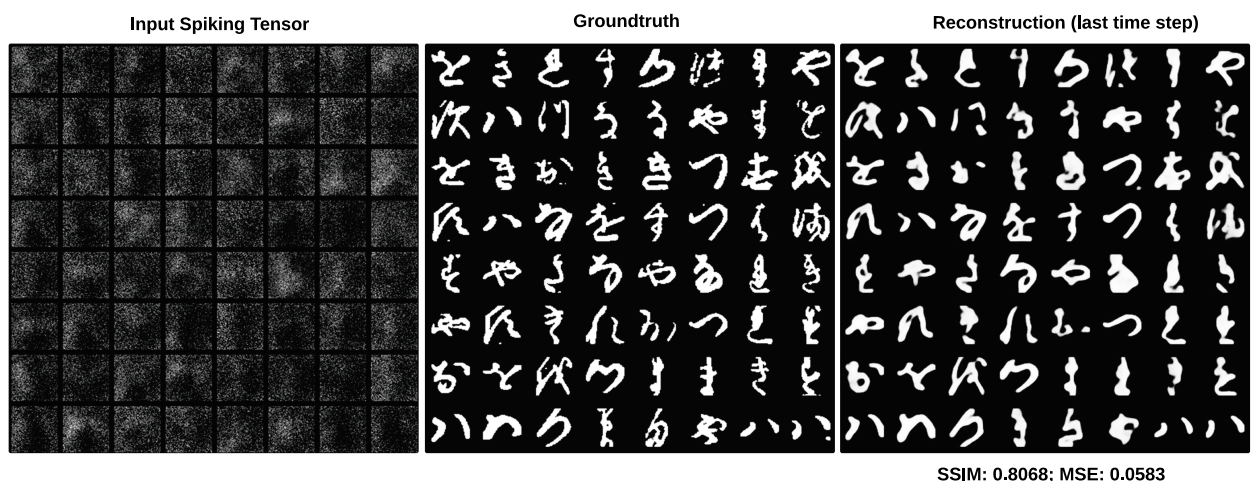

Figure 6. Results of experiments conducted in reflection (bakscattering) geometry using Kanji characters as targets. Each character was dynamically presented—emerging and disappearing once per object—on an E-ink display to mimic optically dynamic but spatially stationary targets . Doubling the optical pathlength through the obscuring phantom rendered the raw output from the DVS camera entirely unrecognizable. Nonetheless, the SNN-based approach produced reconstructed images that were clear and unambiguous. Training and testing used 25,000 and 3,000 characters, respectively. Quantitative evaluation yielded an SSIM of 0.8068 and an MSE of 0.0583

exponentially with the diffusive optical pathlength. The raw output from the DVS camera was entirely unrecognizable. However, applying our SNN-based approach yielded the reconstructed images (right panel) were unambiguous. Here the first 25,000 characters from the full training set and 3,000 characters from the testing set were used for training and testing, respectively.

Note that we specifically chose to use an LED as the light source to better represent real-world applications where incoherent light sources are more commonly used. This adds a challenge to the reflection setup, as it cannot benefit from the high spatial and temporal coherence along with higher energy intensity, that lasers provide. Furthermore, strong surface reflections pose an additional challenge to the camera's dynamic range and increase background noise, reducing the signal-to-noise ratio. Nonetheless, quantitative evaluation across the entire testing dataset yielded an SSIM of 0.8068 and an MSE of 0.0583, suggesting that even in this case of very low incoming information content, useful image reconstruction is possible. Reflection experiments pave the way to potential applications in various fields, such as biomedical imaging and autonomous driving, where reconstructing images from heavily scattered light remains a key challenge.

## *ENERGY CONSUMPTION – COMPARATIVE ESTIMATE BETWEEN SNN AND ANN-BASED MODELS*

We computed the energy consumption of our SNN engine and compared it with a traditional ANN of the same overall architecture (Table I). The computations are categorized into MAC (Multiply-Accumulate) operations, which involve floating-point data processing (including one multiplication and one addition per operation), and AC (Accumulate) operations, which consist solely of additions and are primarily used for processing binarized spiking data in the SNN.

In the SNN design, following the first spiking neuron layer, all subsequent data flows in the form of spikes (binary values: 0 or 1), resulting in AC-dominant operations and hence higher computational efficiency. In contrast, ANN architectures maintain floating-point data throughout the network, making MAC operations predominant. Furthermore, SNNs operate on an event-driven basis, where computations occur only when spikes are generated. This feature, combined with the sparse firing activity of spiking neurons, reduces both computational load and energy consumption.

In a comparative evaluation of our SNN design, we applied inference looping through all test data and monitored the average spiking rate, i.e. evaluated the ACs and MACs for every layer of the SNN as well as the ANN engines[72], respectively, by using the method of Ref 55. To facilitate a head-to-head comparison, LIF neurons in the SNN were replaced with ReLU neurons in the ANN while conserving architecture. The total energy consumption was estimated using the equation below

$$E_{total} = T \times [O_{MAC} \times E_{MAC} + O_{AC} \times E_{AC}] \qquad [7]$$

Where T represents the total number of time steps, $O_{MAC}$, $O_{AC}$ are the number of MAC and AC operations at each time step, respectively. $O_{AC}$ depends on the current spiking rate in each layer. We computed the estimated energy cost in 45nm CMOS chip technology based on previous works[73–75]: the energy cost $E_{MAC}$, $E_{AC}$ for each MAC and AC is 4.6pJ and 0.9pJ respectively[76]. Both SNN and ANN designs require multiple time steps for tracking and reconstruction, but the SNN demonstrates significantly lower energy consumption (7.99mJ, up to 20× lower than the ANN) due to its event-driven computation, sparse firing activity, and reliance on AC operations. Additional details of the calculations are provided as Supplemental Material S1.

| Model | Input Resolution | Params(M) | OPs(G) | ACs(G) | MACs(G) | Energy (mJ) |
|-------|------------------|-----------|--------|--------|---------|-------------|
| SNN   | [18,2,64,64]     | 3.69      | 34.81  | 4.95   | 0.769   | 7.99        |
| ANN   | [18,2,64,64]     | 3.69      | 30.94  | 0      | 30.94   | 142.32      |

Table I. Comparing energy consumption between the SNN and ANN of the same overall architecture, showing the dominance of AC operations in the SNN vs. MAC operations in the ANN, respectively. The last column highlights the energy benefits of the event-driven, sparse firing of our SNN model. Params: total parameters in the network. OPs: total operations. ACs and MACs are total number of MAC and AC operations over all time steps.

## **DISCUSSION**

While neither the use of an event sensing camera nor the design of a spiking neural network engine are new, the synergy inspired by the human vision system has enabled us to construct an end-to-end neuromorphic solution to a challenging optical tracking and object recognition problem in the presence of dense scattering media. As shown in Figure 4-6 and the Supplementary videos, objects which are entirely unrecognizable to naked eye can be brought to a sharp focus by integrating a DVS camera with our dual module multi-stage SNN design. Importantly, the large gain in energy efficiency of this particular SNN design as demonstrated in Table I offers considerably benefit particularly in future mobile/portable applications. Of course, as with any SNN-based approach, the operating assumption here is that the events of interest are sparse. We have not compared our all-neuromorphic method with those possible approaches that deploy frame-based cameras or recent SPAD array cameras as the sensors. A good comparison of the respective camera performance as such is given in this paper[77].

In this paper we have not attempted to compare regular frame-based cameras with the event-sensing approach deployed here. We suggest that a direct comparison is not particularly meaningful given the data requirements and benefits of the spiking neural network models we have developed. While some studies have employed frame-based cameras with artificial neural networks for imaging or tracking through scattering media[8,57], there is limited work combining both functionalities in a fully integrated approach for randomly moving targets demonstrated here in both transmission and reflection. Most work in general has deployed simple scattering media such as ground glasses and rarely quantify the scattering medium in terms of effective optical thickness.

There are a number of improvements which can be envisioned to build upon based on these. On the optical engineering side, for example, the emerging availability of chip-cale semiconductor laser arrays such as those based on vertical cavity surface-emitting laser (VCSEL) arrays offers the possibility for multi-channel extension from the single source in this study[78,79]. Improving the sensitivity, contrast ratio, and noise performance of the event sensing cameras themselves be highly desirable. Adding a second camera to mimic binocular vision, hence achieve depth perception, can provide for an opportunity for 3D imaging as we believe that the added computational burden can be managed by our particular SNN model. Similarly adding a time-of-flight (ToF) capability to the optical system can add significantly further value, provided that DVS-type cameras can be augmented with an ultrafast (< nsec) front-end optoelectronic or electronic shutter (as such, today's VCSEL arrays can readily reach multi-GHz speeds).

On the computational side, our choice of a U-Net-like approach for the core of the SNN as well as the accompanying choice of the surrogate backpropagation method is but one possible way to develop a deep SNN algorithmic strategy. We note that in many SNN designs, multiple time steps are utilized primarily to achieve higher accuracy when processing static input data[75,80–83]. However, this often involves repeated time steps, which increases energy consumption and reduces efficiency. In contrast, when working with dynamic targets and time-series information, the multi-step capability of SNNs becomes a unique advantage. For dynamic scenarios, spike patterns vary across time steps due to motion. By integrating more information—represented by diverse spike patterns correlated with the object's motion—longer temporal sequences allow the network to converge on more accurate reconstructions.

Our network is trained iteratively, step by step, through the full temporal length. This process leverages the memory capabilities of the leaky integrate-and-fire (LIF) neurons and synapse filters, enabling the SNN to effectively learn and utilize spatiotemporal data over time. Temporal memory is particularly critical for processing DVS motion variance. Objects in motion can generate similar spike patterns depending on their direction, speed, and acceleration. The SNN's ability to use its temporal memory to differentiate these patterns ensures accurate tracking and reconstruction of dynamic objects.

Finally we note that to the best of our knowledge, all previously available neuromorphic datasets (also mainly for classification applications) are based on fixed trajectories[84,85]. In contrast to random object movement (such as in this paper), fixed trajectories provide prior knowledge that can simplify algorithm design. However, this can also lead to overfitting, as algorithms may overly rely on the predictable motion patterns rather than generalizing to more complex or dynamic scenarios.

## CONCLUSION

In this paper we have introduced a new neuromorphic optical technique to track and image dynamic targets obscured by turbid media. The objects may be randomly moving or stationary but with time-varying optical contrast. We have engineered an end-to-end optical imaging system by an approach which mimics salient features of the human visual system. The neuromorphic system integrates event-driven optical engineering hardware with a neuromorphic decoding scheme whereby binary spikes carry and are deployed as the sole currency unit of information. More specifically, spatio-temporal changes in light intensity which are collected by a DVS camera are transformed into spikes in real time akin to neural impulses in the retina, then forwarded to custom special-purpose modular neuromorphic computational machinery for image reconstruction. In particular, we have developed an approach to deep learning using a new type of spiking neural network (SNN) model and demonstrate its ability to reconstruct of different types of characters drawn from known datasets.

As such, our study is a proof-of-concept demonstration of an entirely new approach for addressing an important problem of significant interest in pushing optical engineering and computational techniques towards the fundamental limits imposed by the inherently diffusive nature of photon propagation in turbid media. Our work represents the synthesis of powerful camera technology (DVS detectors) with the opportunities provided by neuromorphic computing (SNN model). In leveraging the inherent qualities of deep machine learning, here for the case of SNNs, our study demonstrates the promise these algorithms hold to reconstruct spatial and temporal information in difficult optical domains and paves the way to further development towards applications requiring dynamical object reconstruction in real time.

The results offer broad potential for optically imaging targets of interest or complex phenomena where visibility is entirely obscured by light scattering turbid media. Applications such as orbital-monitoring of wildfires or identification of vehicles moving in dense fog might be examples. We envision the technique of having value also in microscopy of dynamical biological targets where recent progress has been made in computational imaging and adding to the photonic toolkit e.g. in neuroimaging[86]. Given how we used spatially structured light, the neuromorphic approach might find opportunities in free-space optical communication as well as in non-line of sight imaging[87]. With the emphasis on using binary spikes as the unit of information, the end-to-end neuromorphic approach of this paper can be extended to include low-error rate wireless transmission of data prior to decoding[56]. Finally, given the emergence of low-power, low-latency, compact portable neuromorphic hardware such as Intel's Loihi[88] and others, the scheme in this paper could be particularly attractive for those sensor applications in the field where system mobility/portability is of the essence.

## METHODS

### Spiking Neuron: Leak-Integrate-and-Fire (LIF) Model

LIF is selected to be the basic neuron model of our deep SNN. It incorporates time-dependent dynamics and discrete spike outputs, mimicking biological neurons more closely comparing to neurons in ANNs. It maintains an internal state (membrane potential), exhibits a leakage of potential over time, fires spikes when a threshold is reached, followed by a refractory period. The LIF neuron dynamics can be described by:

$$V[t+1] = V[t] + \frac{\Delta t}{\tau_m}\left(-(V[t] - Vrest) + \sum_i w_i x_i[t]\right) \quad [8]$$

Where $V(t)$ is the membrane potential at time t; $\tau_m$ is the membrane time constant, representing how quickly the neuron's membrane potential decays towards the resting potential in the absence of input; Vrest is the resting membrane potential; $w_i$ is the synaptic weight from presynaptic neuron I; $x_i[t]$ is the spike train of presynaptic neuron i at time t; $\Delta t$ is the time step used for numerical integration. This is a discrete-time modeling approximation for computer simulations. The equation [8] can be then described as:

$$V[t+1] = \beta V[t] + \sum_i \overline{w}_i x_i[t] \quad [9]$$

Where $\beta = 1 - \frac{\Delta t}{\tau_m}$, represents the decay parameter; $\overline{w}_i = \frac{\Delta t}{\tau_m} w_i$ is the weight with decay. When membrane potential $V[t]$ crosses a threshold $V_{th}$, the neuron generates a spike, and the membrane potential is reset to resting potential.

$$S[t] = \Phi(V[t] - V_{th}) \quad [10] \qquad \Phi(x) = \begin{cases} 1, & x \geq 0 \\ 0, & x < 0 \end{cases} \quad [10]$$

In contrast, ANN neurons, such as those using the ReLU, Sigmoid or Tanh activation functions, are simplified models, process inputs instantaneously without maintaining an internal state, producing continuous output values instead of discrete spikes without a refractory period. While the simplicity and efficiency of ANN neurons have made them the standard in current deep learning ANN architectures. The biological realism of LIF and other SNN neurons makes them more valuable for neuromorphic computation, gaining advantages in sparsity, energy-efficiency and handling complex neuronal simulations.

*Surrogate Gradient Function*

Gradient needs to be calculated when train a SNN using gradient descent. However, the derivative of the LIF neuron's Heaviside function Φ (x) (equation [10],[11]) returns either +∞ when x = 0, or 0 when x ≠ 0, which prevents learning. This issue is commonly referred to as the "dead neuron problem". To train our SNN network, a surrogate function $\sigma(x) = \frac{1}{\pi} arctan(\frac{\pi}{2} \alpha x) + \frac{1}{2}$ [arctangent (ATan)] is used to replace the Heaviside function Φ (x). ATan function is continuous and differentiable, making it suitable for backpropagation in SNNs. This approach helps to bridge the gap between the biological plausibility of SNNs and the effectiveness of traditional learning methods, allowing SNNs to benefit from gradient-based optimization.

*Data Preprocessing*

Instead of capturing static frames, a DVS camera records changes (events) in incident intensity at each pixel independently and asynchronously (and logarithmically), once a change exceeds a predefined threshold:

$$\Delta L(x,y,t) = L(x,y,t) - L(x,y,t - \Delta t) = \begin{cases} +1 & \text{if } \Delta L(x,y,t) > \theta \\ -1 & \text{if } \Delta L(x,y,t) < \theta \\ 0 & \text{otherwise} \end{cases} \quad [11]$$

where L(x,y,t) is the log intensity of light at pixel coordinates (x,y) at time t; Δt is the interval since the last event at pixel (x,y); θ is a threshold that determines the sensitivity of the sensor to changes in light intensity

The generated binary spike events along the temporal axis can be organized into a stream consisting of $N$ spike events, expressed as a tuple: $\{e_i\}_{i=1}^{N} = \{(x_i, y_i, t_i, p_i)\}$, where p signifies spike polarity (+1 or -1; increasing or decreasing light intensity), x and y are coordinate values of that pixel in sensor's x-y plane, t is time step in microsecond.

The inputs to our SNN model are spike input tensors preprocessed from the raw spike trains. We begin by selecting data region of interest (ROI) to only select spike trains from a certain number of sensor pixels. This is followed by applying a selection of noise filters: activity noise filter, spatial-temporal contrast filter, and antiflicker filter (see details in Supplement) to remove unwanted spikes so as to retain only informative events caused by dynamical variations in the optical scene. Subsequently, the denoised spikes undergo a dimensionality reduction process; we chose here a bio-inspired insect eye method[89] in reducing spatial dimension while preserving essential information. Lastly, the refined spike events are organized through a 3D binning process in both spatial and temporal domains, grouping them as a sequence of event time windows. An input tensor takes the form of $[T, 2, H, W]$ where T is the total number of time steps; the digit 2 includes both positive and negative polarities.

*Training and Evaluation of our SNN Model*

As model targets in the benchtop optical experiments we chose two widely used datasets in computer vision: MNIST (hand-written digits) and K-MNIST (hand-written kanji characters). Each have geometrical features from which more complex objects can be composed. Both sets have 60,000 training and 10,000 testing data. We choose first 10,000 data from its original training dataset, and first 2,000 data from original testing dataset, as our training and testing datasets. This was a pragmatic choice: the reduced size of data can speed up benchtop experiments and computation process, while still sufficiently large to train the model.

The reconstructed image $R\ [T, 1, H, W]$ is obtained from the accumulated membrane potential from the spiking neuron layer of output block, i.e. the last time step in a series of time steps. The corresponding target images are from original datasets after preprocessing (resizing, binarizing, etc), to match the reconstructed image. Target images represent the full features of dynamic objects. In the experiments, a target image $G\ [T, 1, H, W]$ is a gray image at each time step. A combined loss function, comprising mean square error (MSE) and structural similarity index measure (SSIM), is used to quantify the errors between the reconstructed images and target images for both training and evaluation:

$$\boldsymbol{L}(R, G) = a \cdot BCE(R, G) + b \cdot SSIM(R, G) + c \cdot IOU(R, G) \quad [12]$$

The detailed parameters employed in our model are listed in the table shown in Figure 3. We refined our model using the *AdamW* optimizer over 50 epochs on the training dataset. Additionally, we conducted 5 iterations of training and testing, each with a different initialization. We confirmed that our models were adequately trained by analyzing their training and testing loss profiles. The corresponding plot for training and test losses is available in the supplementary. We observed no signs of overfitting as the testing loss consistently decreased alongside the training loss with each iteration.

*Phantom Preparation*

We strictly follow the methods and materials described in references[90,91] to fabricate the optical phantoms used in the benchtop experiments. The primary substance for creating optical phantoms is SiliGlass, a transparent silicone rubber procured from MBFibreglass (PlatSil SiliGlass), which has two components: part A (base material) and part B (hardener). To achieve specific optical properties, a black silicone pigment known as Polycraft Black Silicone Pigment is used as an absorber (diluted in part A at a ratio of 2272:1), and silica microspheres (440345, Sigma-Aldrich) are employed for scattering. Phantom preparation involves mixing part A of SiliGlass with the diluted absorber for absorption and part B with the disaggregated microspheres for scattering. Both mixtures are stirred magnetically and ultrasonicated for 15 minutes each. After combining at a controlled temperature of 21°C, they are stirred for uniformity and poured into a 3D-printed mold. The resulting phantoms, at particular thickness, targeted specific optical properties for this paper.

*DVS Configuration and Synchronization*

We utilized the Prophesee Evaluation Kit 3.0 – VGA as our DVS camera. This camera has a sensor resolution of 640 x 480, with each pixel measuring 15 x 15 μm. It boasts a latency of 220 μs and a dynamic range exceeding 120dB. By default, the nominal contrast threshold for detecting a new event is set to 25%. However, adjusting the sensor's bias can reduce this threshold to 12%[92]. To enhance photon collection, a C/CS mount lens system is mounted above the camera sensor. During our experiments, the lens aperture was maintained at f/2.8 to optimize light entry to the sensor.

The DVS camera is equipped with synchronization input and output capabilities, enabling the detection of external signals for coordinating between multiple devices. Here, a rising signal edge is interpreted as a start signal, while a falling edge signifies a stop signal. Throughout our experiments, the DVS camera remained operational, ensuring uninterrupted event capture. As part of our display procedure for a target, both the e-ink display and DMD (based on the experimental type) sent a 5V TTL square wave signal. This signal's rising edge preceded the display onset, and its falling edge marked the display conclusion. The DVS system recognized and timestamped these synchronization signals alongside events, offering a microsecond resolution. Once all targets were displayed, the accumulated spike trains (events) were retrieved and partitioned into distinct recordings, according to the identified trigger signals. Subsequent to this, each recording underwent preprocessing to filter out noise events and were tagged with the corresponding target identity, readying them for the subsequent training and testing phases.

*SNN Training and Testing Environments*

We set up our training and testing environment with spikingJelly library[93] and Lava framework[94] for neuromorphic computing [44].

**ACKNOWLEDGEMENTS**
The authors wish to thank Quan Zhang and Gary Strangman at MGH for their many insights. We also acknowledge parallel work in our laboratory by Jihun Lee and Ah-Hyoung. Lee. Research supported by grants from Intel Laboratories (CG70982727 2021) and the Office of Naval Research (N00014-25-1-2061)

**AUTHOR CONTRIBUTIONS**
N.Z. and A.N. conceived the project. N.Z. and A.N. designed the experiments and methodology. N.Z. conducted the experiments and designed the computational toolkit. N.Z. and A.N. wrote the manuscript. N.Z. and AN contributed to the data analysis and provided feedback on the manuscript.

**COMPETING INTERESTS**
The authors declare no competing interests.

**DATA AVAILABILITY**
Data underlying the results presented in this paper will be available in a provided GitHub repository. More information may be obtained from the authors upon reasonable request.